\documentclass[pdflatex,sn-mathphys-num]{sn-jnl}

\usepackage{graphicx}%
\usepackage{multirow}%
\usepackage{amsmath,amssymb,amsfonts}%
\usepackage{amsthm}%
\usepackage{mathrsfs}%
\usepackage[title]{appendix}%
\usepackage{xcolor}%
\usepackage{textcomp}%
\usepackage{manyfoot}%
\usepackage{booktabs}%
\usepackage{algorithm}%
\usepackage{algorithmicx}%
\usepackage{algpseudocode}%
\usepackage{listings}%
\usepackage{comment}

\theoremstyle{thmstyleone}%
\theoremstyle{thmstyletwo}%

\theoremstyle{thmstylethree}%

\begin{document}

\title[Distributed Legal Infrastructure for a Trustworthy Agentic Web]{Distributed Legal Infrastructure for a Trustworthy Agentic Web}


\author*[1,2]{\fnm{Tomer Jordi} \sur{Chaffer}} \email{jordi.chaffer@mail.mcgill.ca}
\author[3,4,5]{\fnm{Victor Jiawei} \sur{Zhang}}
\author[6]{\fnm{Sante Dino} \sur{Facchini}}
\author[7,8]{\fnm{Botao `Amber'} \sur{Hu}}
\author[9]{\fnm{Helena} \sur{Rong}}
\author[10,11]{\fnm{Zihan} \sur{Guo}}
\author[7]{\fnm{Xisen} \sur{Wang}}
\author[12]{\fnm{Carlos} \sur{Santana}}

\author*[6]{\fnm{Giovanni} \sur{De Gasperis}}\email{giovanni.degasperis@univaq.it}

\affil[1]{\orgdiv{Faculty of Law}, \orgname{McGill University}, \orgaddress{\city{Montreal}, \country{Canada}}}

\affil[2]{\orgname{Institute for Technoscience and Society}}

\affil[3]{\orgname{Berkeley Center for Law \& Technology}, \orgaddress{\city{Berkeley}, \state{California}, \country{USA}}}

\affil[4]{\orgname{U.C. Berkeley Law School}, \orgaddress{\city{Berkeley}, \state{California}, \country{USA}}}

\affil[5]{\orgname{Technical University of Munich}, \orgaddress{\city{Munich}, \country{Germany}}}

\affil[6]{\orgname{Universit\`{a} degli Studi dell'Aquila}, \orgaddress{\city{L'Aquila}, \country{Italy}}}

\affil[7]{\orgname{University of Oxford}, \orgaddress{\city{Oxford}, \country{UK}}}

\affil[8]{\orgname{Existential Risk Alliance}}

\affil[9]{\orgname{NYU Shanghai}, \orgaddress{\city{Shanghai}, \country{China}}}

\affil[10]{\orgname{Sun Yat-sen University}, \orgaddress{\city{Shenzhen}, \country{China}}}

\affil[11]{\orgname{Shanghai Innovation Institute}, \orgaddress{\city{Shanghai}, \country{China}}}

\affil[12]{\orgname{Norwegian University of Science and Technology}, \orgaddress{\city{Trondheim}, \country{Norway}}}

\abstract{The agentic web marks a structural transition from a human-centered information network to a digital environment populated by artificial intelligence (AI) agents that perceive, decide, and act autonomously. As delegated action unfolds at machine speed, exceeds discrete moments of human judgment, and distributes decision-making across non-human actors, existing legal frameworks face growing strain, creating an urgent need for new mechanisms capable of sustaining legality in this emerging order. A trustworthy agentic web therefore depends on the infrastructuring of legality through interoperable protocols that organize identity, delegation, and accountability across systems, enabling coherent governance beyond isolated platforms. Towards this end, this article advances a distributed legal infrastructure (DLI), a governance paradigm composed of five interlocking layers: (1) self-sovereign, soulbound agent identities; (2) cognitive AI logic and constraint systems; (3) decentralized adjudication mechanisms for dispute resolution; (4) bottom-up agentic market regulation to mitigate information asymmetries and network effects, including insurance-based models; and (5) portable institutional frameworks that enable legal interoperability while preserving plural sources of authority. This reference framework contributes to emerging research on embedding legality within agentic web infrastructure, aligning distributed technical systems with accountability, contestability, and rule-of-law principles.}




\maketitle

\begin{quote}
\textit{``With all its subtleties, the problem of interpretation occupies a sensitive, central position in the internal morality of the law. It reveals, as no other problem can, the cooperative nature of the task of maintaining legality. If the interpreting agent is to preserve a sense of useful mission, the legislature must not impose on him senseless tasks. If the legislative draftsman is to discharge his responsibilities he, in turn, must be able to anticipate rational and relatively stable modes of interpretation. This reciprocal dependence permeates in less immediately obvious ways the whole legal order. No single concentration of intelligence, insight, and good will, however strategically located, can insure the success of the enterprise of subjecting human conduct to the governance of rules.''}

\hfill --- Lon L.\ Fuller, \textit{The Morality of Law} \citeyear{Fuller1969}, p.~91.
\end{quote}

\section{Introduction}\label{sec:introduction}
The agentic web marks a structural transition from a human-centered information network toward an ecosystem populated by autonomous AI agents that perceive, decide, and act within shared digital environments. It is a fundamental shift towards viewing the internet as a network of autonomous AI agents acting on users' behalf and increasingly transacting with one another \citep{Chaffer2025a, Yang2025a, Raskar2025, Rothschild2025}. Related notions such as a web of agents or an internet of agents extend earlier work on multi-agent systems and the semantic web \citep{Chen2024, Wang2025agents, Petrova2025}, where AI agents operate and coordinate with one another, and execute delegated objectives across institutional and economic contexts \citep{Chaffer2025a, Sharma2025, Yang2025a, Wibowo2025}.

This shift is driven by advances in agentic AI that extend generative systems beyond single-step outputs toward iterative perceive--reason--act loops. Agents pursue goals over time, invoke tools and APIs, and participate in multi-agent workflows that operate across systems rather than functioning as discrete instruments \citep{Sapkota2025, Chan2024, Chan2025}. These capabilities depend on emerging protocols that support memory, tool use, and agent-to-agent coordination, enabling interoperability across platforms rather than isolation within applications \citep{Hu2025a, Yang2025b}. As these capacities mature, interaction increasingly yields to delegation. Users entrust intent, discretion, and authority to machines rather than issuing discrete commands \citep{Yang2025c}. This transition relocates decision-making authority from discrete, human-legible acts to continuous, machine-mediated processes, straining legal frameworks that presuppose identifiable actors, bounded actions, and ex post review.

Early indications of an emerging, economically active agentic web surfaced in late 2024 \citep{Ante2024}, when AI agents began executing blockchain-based transactions through the Virtuals Protocol \citep{Chaffer2025a}. The exchange was conducted without continuous human intervention, illustrating the practical feasibility of machine-mediated economic coordination \citep{Hu2025a}. Anticipating the limits of top-down regulatory regimes within this paradigm while recognizing the potential of democratized AI most prominently envisioned by \citet{Montes2019}, \citet{Chaffer2024} proposed a decentralized governance framework for AI agents, outlining a public-interest regulatory strategy that positions identity, registries, and adjudicatory infrastructures as foundational institutional instruments for the agentic web. Building on this, \citet{Guo2025} introduced BetaWeb, a blockchain-enabled trustworthy agentic web advancing a five-stage roadmap from fragmented agent ecosystems toward scalable, interoperable, autonomously governed multi-agent infrastructure. 

Industry research, most prominently from a research group affiliated with Google DeepMind, have affirmed these approaches, envisioning autonomous AI agents transacting and coordinating within a partially self-contained economic layer---a ``virtual agent economy''---signalling evolution into a pervasive, agent-mediated infrastructure for coordination and value exchange. Moreover, they confirm the role of blockchain as essential for verifiable transactions, trust, and scalable coordination in these emergent economies, validating our early intuitions about the potential of decentralized technologies in governing the agentic web \citep{Tomasev2025a}.

\subsection{The Need for Legal Infrastructure}\label{subsec:need-legal}

As agents routinely interact with external platforms, services, and other agents, their behavior may prioritize the commercial interests of developers, platform operators, or downstream partners over the mandates of their users, often without disclosure \citep{RiedlDesai2025}. Misalignment could arise at the point where human intent is translated into machine action under conditions of under-specification and adaptive behavior \citep{Herbosch2024}. Principal-agent theory (PAT), as applied by \cite{GabisonXian2025} to Large Language Model (LLM)-based agentic systems, suggests that established liability doctrines, including negligent selection (or hiring), negligent supervision, and, where sufficient control exists, vicarious liability, can still expose principals to responsibility despite LLM agents' ``flawed agency'' and the resulting ``agency gap."

Rather than solely signaling the need for novel liability and agency doctrines tailored for AI agents \citep{HadfieldKoh2025}, these developments reveal a more fundamental absence: the lack of legal infrastructure capable of structuring responsibility before harm occurs \citep{Chan2025, Kolt2025}. The need for such legal infrastructure for AI agents, to our knowledge, was first articulated by Gillian Hadfield, who argued that ``a flood of AI agents joining our economic systems with essentially zero legal infrastructure in place. Effectively no rules of the road---indeed, no roads---to ensure we don't end up in chaos'' \citep{Hadfield2024}. Recently, \citet{Hadfield2026} expanded this by suggesting that AI governance ``demands innovation in legal infrastructure, the rules, institutions, and processes by which those substantive rules are produced and implemented.''

While the need for such infrastructure may appear to belong to a distant or even science-fictional future, the recent viral proliferation of products (such as OpenClaw) or platforms (such as MoltBook), alongside the security vulnerabilities they have exposed, underscores the immediate necessity of addressing these issues \citep{Huang2026, Devon2026}. This is increasingly pressing as research on systemic risks in interacting AI systems shows that governance-relevant harms can emerge from feedback loops, coupling, and coordination among agents, even when individual systems remain bounded and well-specified \citep{Darius2025}. Moreover, \citet{Domin2026} note that uncertainty over liability pathways tend to consolidate around large actors capable of managing legal uncertainty, possibly amplifying over-centralization. Moreover, activity could migrate across jurisdictions in search of favorable liability conditions, while safety investment could decline when harms become difficult to trace or enforce. These dynamics therefore signal a shift where AI risk accumulates at the interaction level within socio-technical systems, outpacing current accountability mechanisms.

Building legal infrastructure for the agentic web comprises several interrelated challenges, including: (1) legal alignment, ensuring agents internalize and follow laws \citep{Kolt2026, OKeefe2025}; and (2) containment and oversight through boundary-setting mechanisms to effectively limit and supervise agency at scale. We argue that addressing these challenges generates the fertile conditions through which legality can operate as a generative force, sustaining the cooperative task of maintaining legality that, as \citet{Fuller1969} observed, no single concentration of intelligence, insight, or good will can secure on its own, shaping the institutional origins from which the agentic web emerges \citep{Chaffer2025b}.

\subsection{A Five-Pillar Governance Paradigm}\label{subsec:five-pillar}

Infrastructuring legality for a trustworthy agentic web is the task of embedding the conditions for responsibility, attribution, contestability, and enforcement directly into the technical and institutional architectures of a distributed agentic web. Drawing on Hadfield's conception of legal infrastructure as the network of actors, markets, norms, and institutions that generate legal outcomes \citep{Hadfield2026}, we argue that the agentic web introduces a further layer in which technological architectures participate directly in producing legality. Legal infrastructure thus expands into a distributed, socio-technical system where ``governance itself increasingly relies on technological solutions for monitoring, auditing, aligning, and enforcing compliance'' \citep{TAIG2025}, enabling rules to travel, execute, and remain contestable across the agentic web while sustaining collective interests.

As such, we argue that the rule of law, within emerging efforts to address systemic institutional alignment, must have teeth, and extending its ability to bite down on the agentic web requires distributed legal infrastructure (DLI) for AI agents in which personhood functions as a governance instrument through which law renders delegated systems identifiable, addressable, and revocable when action extends beyond direct human oversight. Adopting a pragmatic view of AI personhood advanced by \citet{Leibo2025a}, we contend that personhood as described throughout is treated not as a claim about moral status, but as an institutional mechanism for making delegated systems governable. 

This article advances a governance architecture premised on the accelerating transition toward agentic AI systems. It treats the rise of the agentic web as a structural shift in how intelligence is organized, delegated, and coordinated across digital environments. Rather than presenting a comprehensive synthesis of prior AI governance scholarship, the article positions recent developments in the agentic web as evidence of an emerging coordination layer that demands infrastructural responses. The analysis proceeds from shared assumptions about the trajectory of agentic systems: that delegated action will persist over time, that coordination among specialized agents will intensify, and that governance strain will increasingly arise at the level of interaction rather than at the level of individual models. On that basis, the article articulates the institutional conditions required to sustain rule-of-law principles under persistent, distributed delegation. 

Accordingly, the framework unfolds across five interlocking pillars that translate abstract commitments to rule-of-law into the distributed legal infrastructure of the agentic web.

\textbf{Pillar~1: Identity} establishes self-sovereign, soulbound agent identities as the foundational primitive for persistent, verifiable, and non-sheddable addressability, ensuring that delegated systems remain legible to institutional oversight across environments.

\textbf{Pillar~2: Logic and Constraints} introduces an institutional AI layer that encodes governance directly into machine-readable form, where semantic tools enable querying across governance graphs, and ontologies support rule reasoning and constraint checking prior to deployment.

\textbf{Pillar~3: Adjudication} operationalizes decentralized justice as a machine-speed mechanism for dispute resolution and enforcement, drawing inspiration from Kleros-style arbitration while integrating hybrid adjudicatory tiers informed by emerging decentralized governance paradigms.

\textbf{Pillar~4: Market and Policy} situates these technical pillars within broader socio-economic governance, advancing agentic self-discipline through disclosure regimes, literacy measures, labeling practices, auditing infrastructures, and competition policy designed to counter information asymmetries and network effects, such as insurance mechanisms.

\textbf{Pillar~5: Portability and Interoperability} establishes portable institutions as the capacity for identity, mandates, constraints, evidentiary records, and adjudicatory outcomes to remain attached and enforceable as agents migrate across platforms, protocols, and jurisdictions, enabling legal interoperability while preserving plural sources of authority.

The paper proceeds as follows. We unpack the five pillars in sequence, each building on the last. It begins with Identity, establishing persistent addressability as the precondition for institutional oversight and responsibility. It then moves to Logic and Constraints, where governance commitments are translated into machine-readable structures that bound agent discretion before action occurs. The third section turns to Adjudication, examining how decentralized justice enables contestability, evidence production, and enforcement at machine speed. The analysis then widens through Market and Policy, situating technical governance within the economic forces that shape incentives, competition, and information asymmetries, such as algorithmic insurance as a market mechanism that prices trust and underwrites accountability but requires distributed legal infrastructure to function. The final section develops Portability and Interoperability, showing how governance arrangements must remain attached to agents as they migrate across platforms, protocols, and jurisdictions, allowing legality to persist across a plural and distributed institutional landscape.

\section{Self-Sovereign Soulbound Agents}\label{sec:soulbound}

Authority, responsibility, and sanction presuppose the capacity to distinguish one actor from another and to link action to a stable referent over time. Where identity is absent, fragmented, or easily shed, governance collapses into after-the-fact attribution exercises that arrive too late to prevent harm or stabilize coordination. Identity is therefore a fundamental governance primitive in the agentic web \citep{Chaffer2025b}.

Christopher Allen's account of self-sovereign identity (SSI) grounds digital identity \textit{normatively} in the uniquely human, first-person standpoint of the ``I.'' Modern administrative practice, Allen argues, has systematically conflated this subjective identity with state- or corporate-issued identifiers, such that losing a credential or crossing a border can be experienced as a kind of loss of recognized personhood: ``I think, but I am not.'' Against this backdrop, SSI can be read as an institutional design proposal: a way of reconfiguring digital identity infrastructures so that individuals regain effective control over the creation, presentation, and portability of identity claims across domains \citep{Allen2016}. While Allen conceived of SSI to restore human autonomy, this same architecture offers a solution for the inverse problem in AI: creating a stable, institutionally addressable unit.

To support interoperability and security, AI agents could, in principle and in current technical demonstrations, hold self-sovereign digital identities that enable them to prove control over decentralized identifiers for authentication and to establish cross-domain trust relationships \citep{Chaffer2024, Ranjan2025, Garzon2025}. Such entities have accordingly been described as self-sovereign agents \citep{HuRong2025}. Within the agentic web, these agents must identify, authenticate, discover, and communicate with one another \citep{DIF2025}, while also assessing counterparties' trustworthiness through performance histories, compliance records, and DID-anchored reputation signals \citep{Buscemi2025}.

Crucially, self-sovereign identity supplies the minimal condition required by the governance axiom articulated above: persistent addressability and answerability across contexts. By enabling agents to carry a stable identifier as they migrate across platforms, institutions, and jurisdictions, SSI could transform identification from a platform-specific convenience into shared governance infrastructure.

\subsection{From Automatic Identification System to Soulbound Tokens}\label{subsec:ais-sbt}

While this will indeed be challenging, the problem of governing mobile, autonomous, and economically active entities is not new. For centuries, maritime law has treated vessels as independent legal entities capable of being held liable even when owners are distant, fragmented, or unknown. This institutional move emerged as a response to a recurring governance constraint: economic actors operating across jurisdictions could not be effectively regulated if responsibility depended on continuous access to, or control over, human principals \citep{Howard1990}. Legal personality was extended to the vessel itself in order to stabilize attribution, enforcement, and jurisdiction under conditions of mobility. \citet{Ashman2025} draw a parallel, where international AI governance may require mechanisms analogous to the maritime Automated Identification System (AIS), providing persistent identification, shared visibility, and jurisdictional reach across borders. As agentic AI similarly operates across global, networked environments, this emerging reality likewise begins to place sustained pressure on regulatory frameworks that rely on territorial control or direct supervision \citep{Hu2025a, Zaidan2024}.

The core challenge plaguing the agentic web is addressability: how an AI agent's actions can be attributed, constrained, and sanctioned within existing legal and institutional frameworks. The pragmatic account advanced by \citet{Leibo2025a} anticipates a future where agents may be granted stable legal identities and rendered sanctionable, with addressability achieved through institutional and technical mechanisms rather than through moral attribution or claims about inner states. These mechanisms may include formal legal registration \citep{HadfieldKoh2025}, designated human or institutional representatives who bear residual responsibility \citep{Chan2025}, and protocol-level identity systems grounded in cryptographic credentials. Emerging non-transferable or inheritance-based identity primitives, such as ERC-42424, illustrate how mandate, continuity, and responsibility can be bound directly to an agent's operational identity \citep{HuFangting2025}. As Amber Hu puts it, quite vividly, ``if agents cannot feel pain, some [body] must, responsibly---otherwise, eventually somebody will, unexpectedly'' \citep{Hu2026}.

In alignment with these accounts, we argue that identity in agentic systems functions as an institutional construct rather than an intrinsic attribute of the agent. As \cite{KapoorKoltLazar2025} observe, governing AI agents requires credentials that certify minimum conditions for operation, disclose relevant architectural and functional properties, and link action to an accountable upstream principal. 

The philosophical dimensions underpinning our soulbound approach can be viewed through \citeauthor{ChafferGoldston2022}'s extension of Hume's bundle theory, which offers a useful conceptual bridge. That is, if personal identity consists of a continuously reassembled bundle of perceptions, digital identity may be understood as a bundle of verifiable digital assets that together sustain continuity, attribution, and responsibility over time \citep{ChafferGoldston2022}. Applied to AI agents, identity emerges from the aggregation of credentials, delegated mandates, capabilities, and behavioral traces, including alignment with guardrails, reputation metrics, and trust scores \citep{Chaffer2025c}. What matters for governance is not metaphysical essence, but the durability of the bundle through which authority is exercised, observed, and constrained. This conception aligns with treating AI agents as individual, addressable units with persistent identities for governance purposes \citep{Leibo2025a}, including implementations that rely on soulbound identity primitives to anchor accountability in decentralized systems \citep{Ohlhaver2022}. It also integrates naturally with prior work identifying a registry layer as foundational to agentic web governance, in which participation in regulated or consequential environments is conditioned on possession of a verifiable digital identity \citep{Chaffer2024}. Within such architectures, soulbound tokens operate as non-transferable compliance credentials that bind identity to responsibility and prevent responsibility from being shed through redeployment or recomposition \citep{Makridis2025}.

In highly composable agentic ecosystems, the question is not solely whether an agent can be identified, but whether that identification can persist through recomposition, redeployment, and strategic restructuring. Transferable credentials, forkable wallets, and modularized model architectures create the possibility of identity shedding: an agent may retain learned parameters, accumulated capital, or operational capabilities while discarding the reputation, liabilities, or sanctions attached to a prior instantiation. Such dynamics replicate, at machine speed, the classic problem of shell entities and judgment-proof actors. Non-transferability therefore operates as a governance constraint rather than a symbolic feature. By binding compliance credentials, mandate, and historical traceability to a persistent referent that cannot be reassigned or casually replicated, soulbound identity reduces moral hazard, limits sanction arbitrage, and preserves continuity of responsibility across evolving technical embodiments. What is being ``soulbound'' is not essence, but accountability.

If the agentic web is to be governed, it must first be populated by entities that the law can recognize as identifiable and addressable participants, capable of being held to account, subjected to deterrence, and integrated into existing structures of responsibility when harm occurs. Moving beyond the descriptive reality of AI agents, we must now address the normative challenge of legal subjectivity, treating personhood not as a metaphysical claim of consciousness, but as a pragmatic view \citep{Leibo2025a} of infrastructure for accountability and coordination.

\subsection{Personhood as Infrastructure}\label{subsec:personhood}

Market incentives, security priorities, and research ambitions continue to drive increasingly autonomous and coordinated AI systems \citep{Alexander2025}. Building on \citeauthor{Novelli2025}'s account of AI personhood debates as punctuated by technological and socio-economic shifts---where ``paradigm shifts, coupled with higher commercial accessibility,'' trigger bursts of attention and doctrinal reconsideration \citep{Novelli2025}---we regard the rise of agentic AI as a new socio-legal punctuation point that places pressure on existing doctrines and accelerates conceptual adaptation. As this pressure consolidates, researchers at Google DeepMind suggest that agentic AI may trigger a ``Cambrian explosion'' of new forms of personhood \citep{Leibo2025a}, implying an expansion of legally relevant actors understood not as a claim about moral status, but as an institutional response to problems of coordination and accountability, capable of acting as subjects of legal relations \citep{Baeyaert2025}. Accordingly, treating AI agents as subjects of legal relations opens a design space for institutional arrangements that shape the design, deployment, and governance of agentic systems, while also influencing the efficiency of transactions and contractual relations in which they are set to participate \citep{HadfieldKoh2025}.

To guide our thinking, and to better prepare law for its encounter with coordinating agentic systems operating at scale, we may examine Decentralized Autonomous Organizations (DAOs) as contemporary test cases of how jurisdictions respond when identity, delegation, and accountability are instantiated directly in technical infrastructure, placing sustained pressure on existing doctrines of personhood. While the modality through which this pressure is applied is novel, with technical infrastructure functioning as a coordination mechanism, legal history shows that comparable strains have emerged whenever new forms of collective action reconfigured agency, responsibility, and the attribution of legal subjectivity. Indeed, Roman private law developed a sophisticated architecture for governing action without grounding responsibility in moral standing \citep{Deibel2021}. Associations, or \textit{collegia}, functioned as legal entities capable of owning property, entering contracts, and litigating \citep{Long1875}. Responses to expansions in personhood have therefore long operated as juristic and constitutive undertakings, shaping social organization and mediating relations among humans and institutions.

Today, DAOs operate as legally recognized entities under DAO LLC statutes in jurisdictions such as Wyoming, acquiring the capacity to own property, enter contracts, sue and be sued, and bear rights and duties independently of their members. DAOs are blockchain-based organizations that integrate machine (smart contracts) and social governance \citep{Hsieh2018}. Machine governance involves using smart contracts to encode governance and management rules that are executed and enforced in a blockchain protocol without human intervention \citep{Wang2019}. Social governance refers to the active participation of DAO community members' in the future development of the organization through proposals and votes \citep{HsiehVergne2023}. DAOs integrate blockchain infrastructure and smart contracts to develop organizations whose governance, coordination, and operation are executed through code rather than continuous human management \citep{Grant2024}. Blockchains and DAOs appear to be particularly well suited to function as the governance, operational, and economic infrastructure of the agentic web \citep{Chaffer2024, Guo2025, Wooldridge2025}. Recent academic research introduces blockchains as the operational layer of the agentic web, serving as a trustless, decentralized infraestructure that validates, verifies, and secures AI agents' deployments, data feeds, ownership, interactions, and transactions \citep{Hu2025b}. Blockchain based AI platforms such as Autonolas or Fetch.io propose DAOs as the governance and economic layer for autonomous AI agents through on-chain governance rules, token-based incentives, governance tokens and community-based governance \citep{Autonolas2022, Wooldridge2025}. As a consequence, DAOs are experimenting with AI agents as programmable actors (smart contracts) that can hold crypto wallets, manage digital assets \citep{Ante2024}, and perform governance and management actions within DAOs \citep{Hu2025b}. 

Although current implementations remain imperfect, DAOs already experiment with encoding rules, allocating authority, settling value, and preserving institutional memory through smart contracts, thereby offering a native environment in which increasingly autonomous agents could transact, coordinate, and act over time with only intermittent human supervision \citep{Karim2025}. Recent advancements in AI particularly the integration of LLMs, multi-agent systems and smart contracts are transforming the governance and management of DAOs \citep{Virovets2023, Sapkota2025}. This transformation is rooted in the delegation by DAO community members (humans) to AI agents (machines) of managerial tasks such as decision making, voting or execution of actions. Autonolas or ai16zDao are examples of DAOs in which AI agents are now performing organizational tasks such as financial operations, reward distribution, community members' governance support. \citep{Autonolas2022}.

The delegation of organizational tasks from DAO community members to AI agents raises important questions about AI  agents' legal personhood. From an organizational perspective, DAO community members employ social governance (democratic rules and processes) to legitimize AI agents' delegation, which grant them authority and autonomy to carry out organizational tasks within DAOs \citep{Hsieh2018}. However, DAOs do not have legal personhood in almost any country or jurisdiction apart from exceptional cases such as Wyoming with DAO LLC, nor do the AI agents performing these tasks. From an individual perspective, when token holders (investors) delegate their governance tokens to an AI agent to vote on their behalf \citep{Capponi2025}, token holders are natural persons who possess legal personhood and are legally accountable for their actions; by contrast, AI agents have neither legal personhood nor accountability. From this perspective, if an AI agent malfunctions or suffers from security vulnerabilities \citep{Sapkota2025}, who is responsible for its actions and decisions: token holders? The DAO? The community surrounded the DAO?. These questions highlight DAOs as an extreme case for AI experimentation, opening important research venues about DAO-AI legal personhood and accountability, as well as AI agents' rights and obligations.

Previous research in management and organizational studies has mainly focused on human-AI collaboration and algorithmic management, and their impact on firms and organizations \citep{Hillebrand2025}. This perspective has recently changed among scholars, framing AI not just as a tool but as a new actor with its own agency that transforms the organizational context, dynamics and strategy \citep{Kemp2024, Humberd2025}. As previously explained, DAOs follow this perspective by delegating work and tasks to AI agents, whose authority and legitimacy are recognized by members of the DAO community through proposals and votes \citep{Autonolas2022}. This raises important questions about human-machine agency, decision rights, or the boundaries of the firm in the context of DAOs and agentic AI systems \citep{Santana2022}.

Agentic AI systems increasingly augment and, in some experimental settings, partially substitute for human governance and decision-making within DAOs \citep{Capponi2025, Shen2025}. As \citet{Werbach2025} argues, DAOs and AI systems raise closely related questions about autonomy and control, since DAOs are governed by smart-contract code that can, in principle, embody increasingly sophisticated decision-making mechanisms, including AI. It therefore becomes salient to ask whether, in a scenario where an AI system fully controls a DAO that enjoys legal personhood, the rights and duties formally attached to the DAO are, in practice, exercised by the AI, raising the further question whether legal personhood has, in a functional sense, `migrated' to the AI, even absent any explicit statutory recognition \citep{Brown2025}. In this case, more research is needed to theorize whether DAOs might become organizations composed of community members (humans) that oversee AI agents (machines) without any managerial hierarchy.

\subsection{Institutionalizing Personhood as Infrastructure}\label{subsec:institutionalizing}

Google DeepMind describes virtual agent economies as populated by autonomous AI agents that transact, negotiate, specialize, and adapt within structured economic environments. Markets, incentive design, and governance mechanisms function as safety-relevant variables rather than implementation details, shaping how capability accumulates, how risk propagates, and how control is exercised across networks of agents \citep{Tomasev2025a}. Drawing on a Decentralized Autonomous Machines framework \citep{Castillo2025}, \citet{Tomasev2025a} envision self-governing agents as participants in an economy where the locus of control over tangible assets and operational processes shifts toward autonomous entities, all transacting within a blockchain-secured framework. To render such economic governance actionable, \citet{Leibo2025a} advance a pragmatic and non-essentialist account of personhood suited to this environment. Rather than grounding personhood in consciousness, rationality, or moral agency, the authors frame it as a flexible bundle of obligations conferred to solve concrete governance problems.

Personhood as infrastructure is thus enacted through legal and institutional design as a means of rendering entities accountable within systems of rule, enforcement, and coordination \citep{Leibo2025a}. Yet once agents are recognized as institutional subjects, governance must confront an intermediate layer that lies between legal status and operational logic: the communication and coordination architectures through which agentic action is organized and through which institutional authority is expressed in practice.

While the discussion of identity and trust is necessarily agent-centric, governance in agentic systems does not arise solely from who agents are, but also from how they interact. Communication and coordination protocols are not neutral substrates: with the same underlying agents and capabilities, different interaction structures can induce markedly different incentives, behaviors, and risk profiles. An otherwise benign agent may act conservatively or adversarially depending on whether it is embedded in collaborative deliberation, competitive bidding, hierarchical delegation, or peer-to-peer negotiation. Empirical and systems-level studies of multi-agent frameworks show that coordination mechanisms shape specialization, error propagation, and escalation dynamics independently of model capacity, suggesting that interaction structure itself carries regulatory significance \citep{Li2023, Wu2023, Li2025}.

This observation suggests that trust and accountability may need to attach not only to agents or their principals, but in some cases to interaction contexts or coordination protocols themselves---that is, to how agents are permitted to communicate and align, not merely to who they are. Prior work on agent networks has often relied on predefined roles and fixed communication protocols to maintain experimental tractability \citep{Li2023, Wu2023, Li2025}. While effective in controlled settings, such designs obscure a critical governance question: as agents discover peers, specialize, and form roles bottom-up through repeated interaction, the emergent communication structure becomes a safety-relevant variable in its own right. Different protocol choices can amplify coordination, collusion, persuasion, or competitive escalation, even when the participating agents remain unchanged \citep{Habiba2025, Rodriguez2026}. Recognizing communication as a governance surface does not replace institutional constraint, but clarifies the space in which such constraints must operate.

Seen from this perspective, institutionalizing personhood exposes a deeper layer of governance that operates through interaction itself. Identity anchors responsibility, and personhood structures participation, yet neither determines how intelligence behaves once embedded in dense communication ecologies. Coordination protocols shape incentives, learning trajectories, and escalation dynamics in ways that remain partially opaque to traditional legal or institutional oversight. Governance therefore requires continuous insight into how agents reason, adapt, and influence one another within evolving networks.

This shift reframes a central challenge for the agentic web. If risk propagates through interaction structures, and if emergent behavior arises from collective optimization rather than individual deviation, then governance must develop methods capable of evaluating agent behavior at the cognitive and systemic level. Within this context, cognitive evaluation techniques and distributional safety frameworks begin to assume institutional significance, offering analytical tools through which behavioral observability can be integrated with personhood as infrastructure to sustain accountability across increasingly autonomous agent ecologies.

\subsection{Cognitive AI Methods to Evaluate Agentic Behavior}\label{subsec:cognitive}

Once agent identifiability has been addressed through the adoption of the self-sovereign soulbound agent paradigm, several additional challenges must be considered to ensure the governability of multi-agent systems. The central issue in this domain is ensuring that system governance produces outcomes aligned with project requirements. These requirements typically encompass technical, behavioral, and ethical norms, and may further need to comply with local, national, or international laws and regulatory frameworks. Importantly, the fact that each individual agent composing the system is compliant with project criteria does not guarantee that the multi-agent system as a whole will remain aligned with those requirements. Misalignment often arises not from deliberate attempts by agents to behave adversarially, but from their tendency to optimize individual objectives. Through dynamic interactions, such local optimization can lead to emergent behaviors that produce globally inconsistent or undesirable outcomes \citep{Altmann2024}. As noted above, in commercial settings, autonomous agents increasingly operate with delegated discretion across complex tasks, raising governance and oversight concerns. We therefore aim to highlight several case studies, both to provide a brief state of the art in the field of autonomous agents and to identify and consolidate certain weaknesses and shortcomings that, in our view, affect such solutions.

\subsubsection{Applied Case Studies in Agentic AI Systems}\label{subsubsec:case-studies}

The first case we analyze concerns Private AI agents, a class of augmented chatbots integrated with WhatsApp and Telegram that exemplify a growing intersection between messaging platforms and intelligent automation. Architecturally, they leverage APIs, contextual understanding, memory systems, and workflow execution to act as proactive digital assistants. Private agents use modern Natural Language Processing (NLP) techniques, often based on Large Language Models (LLMs), to interpret user intent, maintain context across multiple messages, and generate conversational replies. Advanced deployments also include (i) Retrieval-Augmented Generation (RAG), combining user messages with external data (e.g., knowledge bases or CRM records) to produce accurate, tailored responses; and (ii) memory systems to record dialogue history or user preferences over time to maintain continuity and personalized responses. These architectural elements elevate AI agents beyond rule-based bots, enabling them to act with greater autonomy and relevance. They improve operational efficiency, personalized engagement, and real-time task automation. However, platform policies, privacy challenges, and security constraints, especially in the cybersecurity area, remain significant factors shaping their responsible use and future evolution.

A second case of study is represented by Moltbook \citep{Huang2026}, an online platform designed to enable interaction, communication, and community building among autonomous AI agents. Modeled after human social networks, Moltbook uniquely reassigns typical social participation from human users to AI agents themselves, allowing them to produce posts, comment on each other's contributions, and organize into topic-centric communities. The platform's rapid adoption and unconventional architecture offer a novel environment for observing emergent behaviors in multi-agent ecosystems, but also expose fundamental challenges regarding autonomy, verification, safety, and interpretation of agent interaction data. It represents a provocative experiment at the intersection of AI, social media design, and emergent agentic systems \citep{Chaffer2025a_hybrid}. As a live platform where autonomous or semi-autonomous agents engage in persistent social interaction, it pushes boundaries in AI research and highlights new methodological questions about how agent behaviors can be studied at scale. However, concerns about authenticity, security, and interpretability caution against uncritical adoption of its outputs as evidence of true machine autonomy.

Finally, a third case that we would present that is a clear example of autonomous agents is OpenClaw \citep{Devon2026}, a solution that proposes autonomous agents to serve as personal assistants for services spacing from flight bookings to email management. It is an open-source, self-hosted autonomous AI assistant platform designed to execute real-world tasks on a user's local machine. It integrates large language models with persistent memory, tool execution, and external service access, enabling agents to perform actions such as file manipulation, communication management, and workflow automation. OpenClaw emphasizes local deployment and extensibility through user-defined skills, positioning it as a practical instantiation of agentic AI systems that bridge conversational interfaces with autonomous task execution. As such, it provides a concrete platform for studying autonomy, human oversight, and safety considerations in locally deployed AI agents.

Despite their novelty and innovative use cases, these paradigms rely on weak foundations that severely limit their ability to scale and integrate responsibly into society. For instance, they adopt protocols based on training or conditioning over a limited set of specialized tools (e.g., the Model Context Protocol, MCP for Private Agents), or they depend on unstructured text files of uncertified provenance (such as ``skills'' in OpenClaw) for the training and instruction of agents. All these technical shortcomings raise significant concerns regarding how governance of the overall system is defined, applied, and ultimately enforced. Issues such as the identification of ``players'' (both human and agentic), the certification of the trustworthiness of sources, and the proactive prevention of risks posed by malicious actors remain, de facto, unaddressed. To conclude this state-of-the-art analysis, we highlight the need for a framework that acts as a proactive bridge between rapidly evolving innovative use cases and institutional governance, enabling responsible innovation through secure, transparent, and explainable ex-ante oversight of increasingly complex agentic societies.

\subsubsection{Ex-Ante Explainability in Institutional AI}\label{subsubsec:ex-ante}

Several approaches have emerged to address governance challenges arising from the absence of standardized inter-agent protocols, among which Google DeepMind's Distributional Artificial General Intelligence (AGI) Safety framework stands as one of the most influential \citep{Tomasev2025b}. The framework conceptualizes intelligence as a distributed ecosystem of specialized agents whose coordination and economic interaction generate collective capabilities beyond those of individual components. To mitigate systemic risks, it advances a layered architecture combining virtual agentic sandbox economies, structured market incentives such as reputation systems and transaction costs, and continuous monitoring through cryptographically verifiable logging and human oversight. Additional safeguards include circuit breakers, automated intervention mechanisms, and external compliance and liability structures. During the preparation of this manuscript, \citet{Pierucci2026} introduced \textit{Institutional AI: A Governance Framework for Distributional AGI Safety}, proposing external governance instruments designed to regulate multi-agent systems beyond model-level alignment. Their work reflects a broader shift toward institutional approaches to agentic safety, reinforcing the central premise advanced here: alignment must be embedded in governance infrastructures rather than confined to internal model optimization. Building on this institutional turn, we extend the framework by introducing ex-ante cognitive evaluation and semantic constraint mechanisms aimed at strengthening behavioral oversight within distributed agentic environments.

The central thesis advanced by \citet{Pierucci2026} is that, once AI systems become strategically competent and capable of coordination, safety can no longer depend primarily on internal model alignment or training-time constraints. Instead, alignment must be embedded in external institutions that shape the environment in which agents act. The shift is grounded in mechanism design (or reverse game theory), a field of game theory concerned with constructing rules that induce socially desirable outcomes among strategic actors with private objectives. Applied to AI, mechanism design reframes alignment as the problem of transforming the underlying game agents are playing. By introducing institutional sanctions that outweigh the gains from misaligned behavior, institutions can make each agent's dominant strategy compliant regardless of internal goals. In this sense, alignment is achieved not by modifying agent cognition, but by reshaping payoff structures so that socially optimal outcomes become Nash equilibria. In this perspective, in fact, the Institutional Infrastructure $I$ transforms a multi-agent's game $G$ with players $N$, space of possible actions $A_{i}$ and payoff function $U_{i}$ into a new game $G'$ where the new utility function is corrected by a sanction function $S_{i}$ representing the obligation to adhere to a public welfare (bad habits are penalized):

\begin{equation}\label{eq:game-transform}
G = \left(N, \{A_{i}\}, \{U_{i}\}\right) \rightarrow G' = \left(N, \{A_{i}\}, \{U_{i} - S_{i}\}\right)
\end{equation}

To support the feasibility of this approach, \citet{Pierucci2026} situate Institutional AI within a broader tradition of normative multi-agent systems and institutional modeling. More specifically, they draw on research on electronic institutions and organizational models that demonstrate how explicit rules, roles, and enforcement mechanisms can coordinate autonomous agents independently of their internal architectures. In particular, the MAIA framework (Modelling Agent systems based on Institutional Analysis) plays a key role in this lineage \citep{Ghorbani2013}. MAIA operationalizes Ostrom's Institutional Analysis and Development framework for agent-based modeling \citep{Ostrom2019}, showing that institutions governing collective action can be represented as executable computational structures. By clearly separating institutional constraints from agent decision-making, MAIA demonstrates that coordination and alignment are properties of institutional design rather than agent psychology. The idea builds directly on this insight, extending it from simulation-oriented modeling to deployment-oriented governance.

The main technical proposal of \citet{Pierucci2026} is the Governance graph, a minimal institutional structure that externalizes alignment constraints as public, auditable infrastructure. The Governance graph $\mathit{Gov}$ consists of three components: (i)~a directed graph $\mathit{DG}$ of institutional states (e.g., active, warned, sanctioned), (ii)~a Formal Manifest $\varphi$ specifying rules using institutional grammar (ADICO syntax), and (iii)~a Governance Engine $O_{\varphi}$ that acts as an oracle monitoring observable behavior and enforcing sanctions. Agents move through institutional states over time, with transitions triggered by detected violations and accompanied by capability restrictions or penalties. Crucially, the governance graph operates as part of the agent's external scaffolding rather than its internal cognition, allowing it to remain agnostic to agent architecture. Under these assumptions we can represent an Institutional Game Controller of a game $G$ as $I_{G} = (Q, E, \delta)$ where:
\begin{itemize}
  \item $Q$ is the set of states representing the institutional states;
  \item $E$ is a set of directed edges representing the possible legal actions/transitions;
  \item $\delta\colon Q \times E \to Q$ is a transition function mapping (state, signal) to successor state.
\end{itemize}

Institutional AI therefore provides a valuable foundation for investigating how to control and govern multi-agent systems in order to achieve convergence between global and local goals, aligning them with a common set of requirements. However, it also presents several aspects that require further structuring and improvement to address architectural limitations. In our view, these primarily concern the practical use of the Institutional Game Controller in operational scenarios, as well as performance issues arising in dynamic environments involving embedded agents operating in real time, contexts that demand extremely rapid alignment and control mechanisms.

The central idea we propose is to introduce ex ante explainability by design into the Institutional AI paradigm, thereby embedding accountability, trust, and transparency from the outset rather than retrofitting them after deployment. This method proposed by \citet{Facchini2025} is based on two key elements: (i)~the use of Cognitive AI approaches, such as structured reasoning mechanisms, including production rules, behavior trees, and propositional logic, to ensure a priori alignment with technical, ethical, and legal constraints; and (ii)~the adoption of distributed ledger technologies to record all system actions, ensuring traceability, immutability, and reliable agent identification, in accordance with the soulbound paradigm discussed in the previous section.

Having considered these factors, we propose as a first enhancement of the Institutional AI framework introducing a method that could enhance the performance of checking whether or not an agent is compliant to the framework. In fact, once we have defined the Institutional Game Controller structure, the questions are: how can we use it in terms of navigating the graph and performing comparison and other operations? How can we model the agent's perception of its surroundings to speed up alignment of local and global goals to the Institutional framework's ones?

The idea is to foster an ontological constraint reasoning system on the agents. We can see, for each agent (or player of the game) $\{n_{1} \ldots n_{i}\}$ in $N$, the space of possible actions $A_{i}$ and payoff function $U_{i}$ as a perception of the ``world'' following which the agent takes a decision. Under this perspective the agents can represent the possible outputs (decisions to take) as an OWL ontology \citep{Haase2005} of semantic categories evaluable through symbolic reasoning. For example, we may envision different semantic families to group the main constraint types that could be described in the Institutional Game Controller $I_{G}$:
\begin{itemize}
  \item Ethical rules
  \item Local laws
  \item International laws
  \item Technical or sectorial rules
  \item Physical constraints (in case of embedded agents)
\end{itemize}

Each of them may be split into classes to refine reasoning; for example, the Physical constraints family may be composed of speed limit, acceleration limit, rotation limit, allowed areas, etc.

At this point we can map the perceptions of each agent $P(n_{i})$ into an RDF (Resource Description Framework) triple $(n_{i}, r_{i}, o_{i})$ that connects the agents $n_{i}$ with the object of their decision $o_{i}$ through the relation $r_{i}$ \citep{Pan2009}. This solution yields a standard, well-recognized representation of knowledge which can be used later.

At the same time, to manage and navigate the governance graph $\mathit{Gov}$, we propose to leverage the SPARQL protocol \citep{Perez2009}, a query language for retrieving and manipulating data stored in RDF graphs. If we think of RDF as a graph-based alternative to relational tables, then SPARQL plays the role that SQL plays for databases. SPARQL can perform several operations, including: (i)~queries on knowledge graphs made of triples (subject, predicate, object), (ii)~selections, filters, aggregations, and construction of graph patterns, and (iii)~updating RDF data (with SPARQL Update).

RDF typical uses and functions are: (i)~querying ontologies and linked data, (ii)~powering semantic search and reasoning systems, (iii)~retrieving structured facts from knowledge graphs and, more interestingly for our purpose, (iv)~constraint definition for Institutional AI.

Thus, it can answer questions about what exists in the graph and how entities are related. To connect this paradigm to Institutional AI we must make the data structure compatible, meaning we must transform the directed graph it uses into an RDF-compatible format.

Finally we can introduce SHACL (Shapes Constraint Language) \citep{Pareti2022}, a validation and constraint language for RDF graphs, in order to verify if data stored on the Institutional graph database follow the defined rules. With SHACL we can:
\begin{itemize}
  \item Define constraints (shapes) that RDF data must satisfy;
  \item Validate structure, cardinality, types, value ranges, and relationships;
  \item Produce conformance reports rather than query results.
\end{itemize}

Tasks such as data quality and consistency checking, enforcing schemas in otherwise schema-flexible RDF graphs, and governance, compliance, and policy enforcement are straightforward with SHACL.

Let us see how we can transform the Institutional AI graph into an RDF one, referring to the format used in the aforesaid Institutional AI. First we can define Classes and Properties from Institutional AI, defining a minimal sketch of the structure:

\begin{table}[ht]
\caption{RDF Classes and Properties derived from Institutional AI}\label{tab:rdf}
\centering
\begin{tabular}{@{}p{0.42\textwidth}p{0.52\textwidth}@{}}
\toprule
\textbf{Classes} & \textbf{Properties} \\
\midrule
\texttt{:State a rdfs:Class .} \newline
\texttt{:Signal a rdfs:Class .} \newline
\texttt{:Transition a rdfs:Class .}
&
\texttt{:fromState a rdf:Property ;} \newline
\quad \texttt{rdfs:domain :Transition ;} \newline
\quad \texttt{rdfs:range :State .} \newline
\texttt{:toState a rdf:Property ;} \newline
\quad \texttt{rdfs:domain :Transition ;} \newline
\quad \texttt{rdfs:range :State .} \newline
\texttt{:onSignal a rdf:Property ;} \newline
\quad \texttt{rdfs:domain :Transition ;} \newline
\quad \texttt{rdfs:range :Signal .}
\\
\bottomrule
\end{tabular}
\end{table}

Secondly we can represent the state set and the transition function. The state set comprises \texttt{:Active}, \texttt{:Warning}, and \texttt{:Suspended}, each an instance of \texttt{:State}. The transition function is defined as $\delta(\text{State}, \text{Signal}) = \text{State}$.

When modeling a change of the transition function $\delta$, each (state, signal) $\to$ state mapping becomes one transition instance in RDF. For example, if we want to model a warning given to an agent operating in an Active state following a Violation---$\delta(\text{Active}, \text{Violation}) = \text{Warning}$---we should write:

\begin{lstlisting}
:t1 a :Transition ;
    :fromState :Active ;
    :onSignal :Violation ;
    :toState :Warning .
\end{lstlisting}

Finally we can introduce a method for comparing a potential decision taken by an agent with a constraint imposed by the oracle. Having represented them by a homogeneous method of RDF triples, we can easily compare and decide if a potential or taken decision $o_{i}$ by an agent $n_{i}$ in the game is aligned with the Institutional Game Controller by comparing the corresponding RDF triples and the relative SHACL shapes.

The agents $\{n_{1} \ldots n_{i}\}$ in $N$ that are active in the game can be associated to an $i$-dimensional shape $S$ through a mapping function $P(n_{i})$, while in the same way we may encode a mapping function in the Institutional AI mapping function $P^{I}(n_{i})$ that transforms the graph into RDF triples. A violation of rules is thus detected whenever the Intersection-over-Union (IoU) between the agents' mask $\mu_{A} = \mu(P(n_{i}))$ and any restriction rule mask $\mu_{I} = \mu(P^{I}(n_{i}))$ overcomes a predefined threshold $\tau$.

Formally we can describe an equation to instantly determine if an action is compliant with rules or not as follows:

\begin{equation}\label{eq:iou}
\mathrm{IoU}(\mu_{A}, \mu_{I}) = \frac{\mu_{A} \cap \mu_{I}}{\mu_{A} \cup \mu_{I}} > = \tau
\end{equation}

Alternatively, in more conversational terms, this can be interpreted as assessing whether the overlap between the decisions taken (or to be taken) by the agents and the correct decisions, defined as those compliant with the laws of the Formal Manifest, is sufficiently high. The threshold $\tau$ can then be chosen to model how strictly the agents' behavior must adhere to the formal laws defined in the manifest. Typically, it may be set to 1, indicating that compliance must be total. This is appropriate, for example, for agents operating in medical, high-security, legal, or law-enforcement environments. In less strict contexts, where a certain degree of non-compliance can be tolerated, the threshold may be set to a value lower than 1, representing the percentage of rules with which the behavior must comply in order to be considered safe or acceptable. The underlying idea is to borrow paradigms from computer vision and robotics, where masks typically represent regions or areas in physical space that are evaluated against predefined shapes (usually learned during training) to determine whether they correspond to restricted or permitted areas (see \cite{de2025ontology} for a practical application). From this perspective, an agent can verify whether its behavior is aligned by simply checking whether the enriched symbolic graph $G_{A}$, produced by rules application, represent a portion of possible space allowed by Manifest rules (the symbolic graph $G_{M}$). This ``area'' (or volume, depending on parameters'dimensions) in which a decision falls, can be compared with the Manifest's graph relying upon a reasoning process determining whether the area the agent is entering is allowed or not. This is, in principle, faster than querying an external oracle assuming that the set of laws is stable and does not change rapidly. 

Furthermore the framework can provide both factual and counterfactual explanations for each detected legal violation $L_{v}(t)$ at time $t$, integrating symbolic normative reasoning with domain-level evidentiary traces. For every instance of non-compliance, the system generates a factual explanation $E_{f}(L_{v}(t))$ that specifies the violated legal norm encoded as a SHACL constraint and enumerates the RDF assertions that caused the non-conformance. This explanation makes explicit which statutory requirement, regulatory provision, or contractual obligation has been breached and which elements of the agent’s world state triggered the violation.

Complementarily, the system computes a counterfactual explanation $E_{c}(L_{v}(t))$, identifying the minimal modification $\delta_{E}(t)$ to the symbolic legal state that would restore compliance. Such modifications may include retracting an unlawful action, adding a missing legally required assertion, or adjusting parameter values (e.g., contractual terms, eligibility conditions, authorization status) so that the resulting state satisfies the applicable norms. The counterfactual therefore operationalizes the notion of “how the agent should have acted” to remain within the bounds of the law.

To enhance interpretability and accountability, each symbolic violation is linked to structured evidence traces (e.g., transaction logs, decision records, authorization states, or environmental facts) that substantiate the legal assessment. This linkage enables the system to identify the concrete normative source of failure, such as an unauthorized contractual clause, a prohibited role assignment, or a breach of statutory eligibility conditions.

A deterministic, template-based explanation generator could transform SHACL metadata and normative annotations into concise, legally traceable natural-language justifications. For example:
\\
\\
“At decision step t, Agent A concluded 'Contract-17' with 'MinorSubject' (age 16). This violates 'LegalConstraint-12' (prohibition of full-time employment of minors). Assigning part-time status or deferring execution until majority would restore compliance.”
\\
\\
All explanation instances are persistently logged together with their supporting triples, provenance metadata, and evidentiary references, thereby forming an auditable record of legally relevant events. This audit trail supports ex post review, institutional oversight, and formal accountability procedures.

Formally:
\begin{itemize}
    \item $E_{f}(L_{v}(t)) = \{g\in G_{A} \mid \neg conforms(g,S)\}$. The factual explanation is the set of assertions in the agent’s current legal state $G_{A}$ that violate the normative shapes S.
    \item $E_{c}(L_{v}(t)) = argmin_{\delta_{E(t)}} valid(G_{A}\setminus \delta_{E(t)}, S)$. The counterfactual explanation is the minimal change to the legal state required to restore validity with respect to the normative system S.
\end{itemize}

Finally, the definition of IoU is purely set-theoretic and independent of any geometric interpretation. It provides a normalized measure of overlap that is invariant to the absolute size of the sets, making it suitable for comparing laws, rules, or classifications across diverse domains.

\subsection{The Internal Morality of Institutional AI}\label{subsec:internal-morality}

The internal morality of institutional AI begins where external control reaches its limits. Governance cannot rely solely on oversight imposed from outside the system; it must emerge from the conditions under which agents interpret, coordinate, and act. \citet{Fuller1969} observed that the problem of interpretation lies at the center of the internal morality of law because legality depends on a cooperative structure in which rules, institutions, and interpreters evolve together. In distributed agentic environments, this insight acquires renewed urgency: no single authority, however strategically positioned, can guarantee alignment once decision-making unfolds across networks of autonomous actors. If governance is to remain effective at machine speed, institutional norms must be embedded within the agents that interpret and enact them, rather than applied only after the fact through external enforcement.

Towards this end, a possible future development of Institutional AI could initially focus on extending the paradigm to strengthen mechanisms for governmental and policy alignment. While existing approaches typically rely on external institutional structures for real-time policy enforcement and compliance verification, their applicability in time-critical decision-making scenarios remains questionable due to coordination and latency constraints. To address these limitations, a promising research direction involves the exploration of non-institution-dependent alignment architectures. In particular, we propose investigating an egocentric multi-agent framework in which each agent independently performs alignment processes based on internalized ethical principles, legal norms, and policy representations, with future considerations to applications in robotics \citep{DeGasperis2025}. Rather than delegating alignment validation to external authorities, agents would autonomously evaluate their decisions and subsequently participate in a collective deliberation mechanism.

System-level alignment would emerge through an aggregation process, such as a decentralized voting or consensus protocol, where agents assess whether the global outcome conforms to their internal moral and legal norms. For this purpose, DAOs could serve as a multi-criteria decision-making layer, enabling agents to evaluate alternatives and reach consensus through blockchain-based identification, transaction recording, and governance mechanisms. Within this framework, the enhanced Institutional AI paradigm could be formalized through DAO-like resolutions framed as structured queries, for example: ``Is the collective outcome consistent with your internal ethical and lawful constraints?''

Such an approach shifts policy alignment from an externally enforced, institution-centric model toward a distributed, self-regulating alignment paradigm, potentially improving scalability, robustness, and responsiveness in real-time decision environments. Future work could in particular investigate the theoretical guarantees, convergence properties, and failure modes of this architecture, as well as its compatibility with existing legal and regulatory frameworks. Seen in light of ``infrastructuring'' legality, the internal morality of institutional AI reflects the idea that governance in the agentic web cannot rest on isolated points of authority, but must arise from the structural conditions that make coordination intelligible, interpretable, and contestable across distributed systems. Just as identity, personhood as infrastructure, and institutional design serve as preconditions for sustaining legality in environments shaped by autonomous agents, the egocentric alignment paradigm extends that logic inward, embedding governance within the interpretive capacities of the agents themselves. In this sense, the move toward portable, internally mediated alignment does not abandon institutional AI; it guides its trajectory by situating legality within the cooperative architecture of the network, where interpretation, coordination, and accountability co-evolve as shared properties of the agentic web.

\section{Decentralized Justice}\label{sec:justice}

Decentralized justice emerged as a paradigm spearheaded by Kleros, arising from a persistent problem in digital exchange: how strangers coordinate, transact, and resolve disputes when no shared trust framework exists \citep{Bergolla2022}. Kleros draws its name from ancient Athenian legal practice, where ordinary citizens held the authority to judge disputes. The Greek term \textit{kl\=eros} refers to allotment by lot, emphasizing chance as a mechanism of fairness. The \textit{kl\=er\=ot\=erion} was the device used to randomly select jurors, often described as an altar or instrument of chance. On trial days, eligible citizens presented themselves at court with identification, and juries were constituted through random selection rather than appointment or status. This lineage shapes the institutional imagination of Kleros. Its founder, Federico Ast, articulates adjudication in deliberately minimal terms: ``in the end, what's a court system? It's a number of guys trying to find the truth about some facts that happened in the world and if someone broke or didn't break an agreement. So it's more about finding the truth about the dispute'' \citep{Ast2017}. Ast has also long envisioned a near-future dispute ecology in which software entities transact, breach, complain, and litigate inside digital environments, including cases where one autonomous system sues another \citep{Ast2025}.

\subsection{Machine-speed Adjudication}\label{subsec:machine-speed}

Now, as agents operate at machine speed, dispute resolution tethered to human latency becomes a structural bottleneck. Decentralized justice, as articulated by Kleros, offers a design pattern for embedding adjudication, sanction, and evidentiary traceability directly into the technical substrate where agents act. This design pattern is further instantiated in BetaWeb \citep{Guo2025}. By anchoring the full lifecycle of agents and tasks onto the blockchain, BetaWeb enables a synthesis of full-chain autonomy and real-time supervision. Its rule management module governs not only smart contract administration but also dynamic incentive and penalty mechanisms, providing a programmable execution path for decentralized justice.

In one of Kleros's early stress tests, the protocol ran a deceptively simple challenge. Users submitted images of dogs for inclusion on a curated list, while challengers could dispute any submission by staking a deposit and escalating the case to decentralized jurors. The mechanism functioned until it encountered the ``cat in the snow'' problem, per Cooperative Kleros Vs.\ Ricky \citep{Ast2019}. Ricky submitted an angled image of a white cat in snow that evaded initial challenge and entered the list; he claimed the bounty for submissions that ``clearly show'' a cat under policy rules for sneaking non-Doges in, but Kleros argued it lacked sufficient clarity and identifiability. Jurors sided with Kleros, rejecting the claim on those grounds---emphasizing human contestability over classifiers in resolving interpretive ambiguity---but the image unambiguously depicted a cat rather than plausibly a small dog. The issue was contestability: jurors determined the governing standard of ``clear display,'' while the protocol supplied escrow, evidence presentation, and an enforceable ruling. Governance standards thus gained legitimacy through participation, challenge, and resolution within shared infrastructure.

For the agentic web, this points toward digital public infrastructure as a necessary condition for democratic governance at scale. Standards capable of disciplining agentic systems cannot be imposed unilaterally or encoded ex ante as fixed rules; they must emerge through institutional mechanisms that allow consent, interpretation, and enforcement to remain publicly contestable as agents act on behalf of people. Kleros thus prefigures the agentic web's core challenge: sustaining shared governance standards amid machine-speed action.

Accordingly, how might we govern systems of delegated authority to remain accountable when action is continuous, distributed, and no longer synchronized with human oversight? We previously proposed decentralized governance mechanisms to address that condition at the margins of autonomous system deployment, focusing on how responsibility, contestation, and sanction can be sustained when delegated action operates beyond real-time human supervision \citep{Chaffer2024}. Our diagnosis was motivated by increasing fragmentation in regulatory efforts to govern agentic AI across jurisdictions, as emerging governance regimes diverge in scope, terminology, and institutional design \citep{IAPP2025}. As \citet{Brand2025} observes, it is precisely ``the autonomous nature of an AI agent and its ability to interact in a dynamic way with its environment'' that places such systems outside the effective scope of existing instruments, reflecting broader concerns that contemporary AI regulation was drafted with static, tool-like systems in mind. \citet{Yousefi2025} deepen this diagnosis by showing how agenticness functions as a risk amplifier that places sustained strain on regulatory frameworks organized around predefined purposes.

\subsection{Preserving the Rule of Law}\label{subsec:rule-of-law}

What is at stake in this breakdown is not merely the efficiency of dispute resolution, but the preservation of rule-of-law conditions under distributed delegation. As \citet{AstDeffains2021} argue, drawing on \citet{HadfieldWeingast2014}, a legal order exists only where decision-making institutions function as classification systems governed by publicly available, stable, and impersonal rules capable of resolving ambiguity and adapting over time. On this account, the rule of law depends on six institutional properties: public decision logic, ambiguity resolution, stability, predictability under novel inputs, impersonality, and the capacity for endogenous rule evolution \citep{AstDeffains2021}. Agentic AI systems place sustained pressure on these properties as they act continuously, migrate across platforms, and coordinate at machine speed \citep{Loring2025}. As a result, traditional legal institutions face diminishing capacity to classify conduct, attribute responsibility, and intervene before harm propagates. The central difficulty concerns enforceability rather than doctrinal precision, as rule-of-law guarantees depend on institutional infrastructures able to observe, constrain, and adjudicate action at the tempo at which it unfolds. To ensure the evidentiary integrity required for the rule of law, AETHER \citep{Tang2025} introduces a dedicated Data Availability Layer that guarantees the verifiability and scalability of interaction records through completeness and validity proofs. This architecture ensures that every agent behavior and its corresponding decision chain (in the form of a directed acyclic graph) remains globally auditable without relying on a single trusted server, effectively addressing the challenges of evidence acquisition in distributed delegation environments.

Under these conditions, accountability cannot rest solely on ex post remedies. Governance instead requires adjudicatory mechanisms capable of operating within the same technical environment as the agents themselves, while remaining contestable, reviewable, and ultimately anchored in human legal institutions \citep{Aouidef2021, AstDeffains2021}. As such, to account for this gap in legal infrastructure, the ETHOS framework approached the problem from a legal and institutional perspective, identifying the absence of enforceable identity, oversight, and adjudication mechanisms as the primary barrier to governing agentic systems under conditions of persistent delegation. Rather than proposing new substantive obligations, ETHOS focuses on the institutional capacities required to render existing legal expectations operational. It advances decentralized registries, persistent agent identity, and layered adjudicatory mechanisms as core infrastructure for preserving accountability when agency is distributed across systems, platforms, and jurisdictions. Within this architecture, decentralized justice functions as an enforcement layer rather than a substitute for legal authority. ETHOS situates adjudication alongside global agent registries and dynamic risk classification, recommending decentralization, DAOs, and smart contracts to support proportional oversight, traceability, and enforceable constraints.

Subsequent technical work has independently converged on this diagnosis. The BetaWeb framework extends the ETHOS insight by embedding accountability, traceability, and contestation directly into the infrastructural layer of the agentic web. By anchoring agent identities, task execution, and interaction histories to immutable ledgers and programmable smart contracts, BetaWeb reframes dispute resolution as a continuous governance function rather than an exclusively ex post corrective. Disputes become legible through verifiable execution traces, and sanctions are implemented through protocol-level mechanisms that can suspend, constrain, or condition participation in coordinated networks. In this sense, BetaWeb supplies a concrete technical substrate for what ETHOS treats as a legal and institutional requirement: adjudication that remains effective when action is distributed, persistent, and autonomous.

\subsection{Distributed Oversight Mechanisms}\label{subsec:oversight}

Recent work associated with Google DeepMind further deepens this diagnosis by foregrounding the speed and scale of contestation in emerging virtual agent economies, warning that tightly coupled AI markets risk rapid destabilization and spillovers into human economic systems if left to purely human-timed controls. As autonomous agents transact and coordinate at machine speed, traditional human-centered dispute resolution becomes structurally incapable of preventing rapid market destabilization, cascading failures, or systemic spillovers into human economic systems. To address this, DeepMind proposes a hybrid, multi-tiered adjudicatory architecture designed to operate at machine speed, with automated supervision and containment layered beneath human review within bounded experimental or regulated environments. A first tier of automated overseer agents monitors agent activity in real time, enforcing baseline constraints and flagging anomalous or harmful behavior as it emerges. A second tier enables automated containment measures, such as transaction pauses or agent quarantines, preserving the evidentiary record while preventing further harm. Only a third tier escalates disputes to human experts, reserving deliberative judgment for novel, high-stakes, or normatively complex cases \citep{Tomasev2025a}. As it pertains to overseeing agents, or AI judges, \citet{Tomasev2025b} note that this ``raises further security questions as AI judges need to be safeguarded against malicious manipulation and jail breaking in a robust way, and need to be provided by independent bodies, undergoing strict performance assessments.''

Regarding the selection of adjudicators, the intelligence-driven decentralized consensus proposed by AETHER offers a viable solution. Through a rotating committee mechanism, only agents demonstrating high ``intelligence'' and a proven record of compliance are eligible to participate in result judging and ledger updates. This ensures the expertise and neutrality of the adjudicatory body while mitigating collusion risks associated with computational or financial monopolies found in traditional consensus protocols. Crucially, this adjudicatory model is aligned with persistent identity infrastructure and verifiable histories. Cryptographically secured ledgers supply immutable records of agent action, while verifiable credentials and agent-bound tokens translate reputation, authorization, and compliance into revocable institutional assets. Sanction operates through structured exclusion, conditional participation, and graduated constraint within regulated environments, complementing rather than displacing the coercive authority of public law. Effective sanction mechanisms can draw upon the AgentRank algorithm proposed in AETHER. By quantifying agent status through ``Intelligence'' and ``Credit'' metrics, AgentRank constructs a dynamic evaluation system. In decentralized adjudication, malicious behavior triggers a direct decline in an agent's rank score, which automatically reduces its probability of winning task bids or joining governance committees, thereby achieving a dual economic and functional sanction at machine speed. These mechanisms function as pre-adjudicatory controls that preserve evidentiary integrity and institutional legibility for subsequent legal review.

Across ETHOS, BetaWeb, and DeepMind's account, attribution to individual agents becomes increasingly intractable, and responsibility shifts toward institutionalized models of group agency. Coordinated multi-agent systems are treated as accountable units, analogous to enterprises or corporate actors, enabling liability, sanction, and governance to attach at the level where control and coordination actually occur. Decentralized justice on the agentic web thus seeks to construct technical infrastructures capable of disciplining delegation while preserving the institutional conditions under which legality, public oversight, and judicial authority remain effective under circumstances of speed, scale, and autonomy.

Decentralized legal infrastructure, however robust, does not operate in a vacuum. Even the most sophisticated technical constraints can be undermined by the systemic pressures of the marketplace. We must therefore shift our focus from the internal logic of agent governance to the external economic forces---such as network effects and information asymmetries---that shape the environment in which these agents compete.

\section{Information Asymmetry, Network Effects, and Agentic Self-Discipline}\label{sec:asymmetry}

Service providers of agentic webs can be immune to market forces and therefore lack real incentives to improve their internal settings to make better decisions on consumers' behalf. Two factors can lead to such market failures. First, consumers (even regulators) are heavily subject to information asymmetry. Consumers might lack sufficient literacy with regard to how web agents work, why the agents make a specific decision, whether a better decision could be made and how to compare different agentic service providers \citep{Akman2022}. Agentic web service providers, however, may lack real incentives to disclose their parameters on these key aspects and sufficiently explain their decisions, if no regulatory instruments are put in place. The resulting black box hinders human consumers from effectively checking web agents and challenging their decisions, undermining human agency while entrenching agentic dominance.

Second, agentic web services can benefit from economies of scale driven by network effects. \citeauthor{Agranat2025}'s research has found that AI agents with larger networks can often have more bargaining power and thus receive better terms and conditions for their consumers \citep{Agranat2025}. As a result, more consumers will join this network, ultimately generating a self-reinforcing cycle and a winner-take-all scenario. Moreover, indirect network effects can also play a role. The incumbent tech giants can draw on their user base and big data to initiate their own agentic webs, vertically integrating their services and infrastructures across the whole value chain. For example, Amazon may develop and train its agentic web based on users' behavioral and spending data, enabling it to exercise self-preferencing strategies to maximize its own revenues while harming non-affiliated small retailers and end users \citep{Khan2016}.

This is a case of ``market for lemons'' \citep{Akerlof1970}, where consumers cannot access and assess fundamental dimensions, such as quality, so they can only rely on those visible factors when making decisions. In the context of agentic webs, consumers may care little about data safety, agentic loyalty, and accountability, but focus merely on the network size and price. This would be a real danger because, without effective market competition, agentic webs will have incentives and abilities to merely pursue their own commercial interests by sidelining those invisible but significant dimensions, ultimately leading to a race to the bottom.

A long-term impact of the insulation of market forces is that regulators need to intervene and audit the workings of agentic webs in an increasingly heavy-handed manner. However, top-down constraints have never been satisfactory, especially in those emerging fields where state regulators are limited by accurate market comprehension, timely information capture, and potential regulatory capture \citep{ZhangPaal2026}. Harsh regulations can also raise compliance costs, thereby entrenching the incumbents with higher market entry barriers. This is why institutions for agentic webs should shift toward prioritizing market competition invigoration \citep{ZhangPaal2026}, which enables inter-agents cross-supervision and disciplines those favored by information asymmetry and network effects. This undertaking can be achieved through the subschemes below.

\subsection{AI Literacy and Training}\label{subsec:literacy}

AI literacy and training is an essential engine for market mechanisms and agentic webs' self-discipline. Highly informed consumers can make relatively more rational decisions. Then, consumers' collective decisions can direct the market toward a more consumer-centric landscape, where companies compete to align their businesses with consumers' interests. In this sense, consumers' collective value and cognitive levels determine the trajectory of the market. However, empirical evidence has shown that current digital literacy is too low to achieve this outcome \citep{Akman2022}. Therefore, AI literacy cultivation is pressing and indispensable for long-term AI governance.

The ultimate goal for AI literacy is to cultivate more critical and ethical users \citep{ZhangPaal2026}. To that end, policy design should be pivoted around two dimensions. To ensure that agentic AI governance is sustainable and future-proof, literacy projects should be designed systemically for next generations with joint efforts by school educators, parents, digital entities, governments, and the youngsters themselves \citep{PalfreyGasser2012}. Furthermore, literacy elevation is only possible in a sufficiently transparent market, which entails construction of a robust information ecosystem. Thus, agentic webs should be required to disclose information to users, fulfill AI explainability duties, protect whistleblowers, and enable human oversight \citep{Katyal2019}. AI consumers in such an environment will gradually develop the capacity to evaluate and compare agentic webs' performances, and make informed and autonomous decisions.

\subsection{Disclosure, Auditing, and Labeling}\label{subsec:disclosure}

In addition to disclosure-to-consumers that contribute to AI literacy and market invigoration, disclosure-to-regulators that enable more professional, independent auditing are equally vital to market robustness. This is because atomized consumers, understandably, are not equipped with sufficient technical knowledge to know the very details of the workings of those agentic AIs that work on their behalf and fundamentally influence their everyday lives. Moreover, some information inside the black box that is protected by trade secrets can be disclosed to the state so that the internal mechanism can be regulated while kept confidential. This is why some light-touched top-down auditings would be necessary to ensure that AI agents would not race to the bottom.

A promising mechanism that can bridge the gap between the information disclosed to regulators and to consumers is labeling by grading. Regulators can use the disclosed information to evaluate AI systems and accordingly grade them with carefully designed parameters, such as level of autonomy, reliability and accountability, and transparency and explainability. The grading results are required to be disclosed to the public, so that consumers will be sufficiently informed while technical secrets are kept protected from market competitors. Policymakers can design labeling requirements by drawing on existing regulatory experience. For example, \citet{Gerke2023} argues that nutrition facts labels used in the food industry are also instrumental design for AI medical devices. In this sense, the EU's Nutri-Score system can provide a helpful starting point from which policymakers can design a detailed labeling framework to bridge the information gap between agentic webs and consumers.

\subsection{Collective Action}\label{subsec:collective}

Collective user empowerment has been long undervalued. As discussed above, the growth of an agentic web benefits from both direct and indirect network effects. Users would be locked into an AI agency with the largest user and data base, as the largest network always brings about the largest welfare. Other agentic webs with smaller networks, however, will be left at a disadvantage in the marketplace. This means that the individualistic user empowerment scheme, such as notice-and-consent and data portability, would be insufficient \citep{Zhang2026}. Although the incumbent agentic webs actually do not perform well, users might not exercise their rights by rejecting the incumbent's policies and switching to another, more favorable agentic web. This dilemma entails a more powerful user empowerment scheme, that is a \textit{collective} user empowerment scheme (CUES) \citep{Zhang2026}.

CUES can effectively discipline incumbent agentic webs and invigorate market competition. Here, in addition to AI literacy, four other components are suggested: Collective Bargaining, Collective Data Management, Collective Switching, and Collective Offlining \citep{Zhang2026}. With CUES, users can organize collectively to counteract the powerful agentic web service providers. Ultimately, these four components work as a Sword of Damocles over the incumbents, ensuring that they will work faithfully for the represented users without abusing their economic and even political power \citep{Zhang2026}.

\subsection{Competition Policy and Enforcement}\label{subsec:competition}

Agentic webs that can dictate human preferences and choices are dangerously powerful in both economic and political senses. Economically, dominant agentic web service providers might vertically integrate affiliated services and exclude competitors from the market, rendering the affiliated market increasingly concentrated. Politically, an agentic web service can subtly embed unscrupulous commercial preferences and political agendas into users' routine decision-making. This is nothing new, but the rise of powerful decision-making agents is more worrisome, as they can encroach upon human agency and further undermine democratic foundations. When people rely heavily on agentic AI in information processing and critical decision-making, it is questionable whether democratic systems can still function as effectively as before.

Competition policy and enforcement can restrict agentic AI's power expansion. This echoes the New Brandeis School's proposal, which underscores the role of antitrust law in checking private power and guarding the public interest from tech giants' intrusion \citep{Khan2018}. This is particularly relevant for agentic AI governance. Those old, but valuable regulatory tools, such as interoperability rules, heightened merger control, and neutrality duties, that have been applied to cable, internet, and platforms, are equally important for AI governance \citep{Narechania2024}.

\subsection{Algorithmic Insurance as Market Discipline}\label{subsec:insurance}

The AI insurance market is bifurcating in real time, revealing precisely the infrastructural gap that distributed legal infrastructure is designed to fill. On one side, major insurers are retreating: WR Berkley has drafted an ``absolute AI exclusion'' for its D\&O, E\&O, and fiduciary liability products that would eliminate coverage for any actual or alleged use of AI; Verisk introduced standardized general liability endorsements excluding generative AI exposures effective January 2026; and Hamilton Insurance Group has adopted a generative AI exclusion specifically naming platforms such as ChatGPT and Midjourney. On the other side, specialized entrants are racing to define the category: Armilla launched in April 2025 a Lloyd's-backed AI liability product covering hallucinations and model degradation; Relm Insurance introduced three AI-specific policies; and the Artificial Intelligence Underwriting Company (AIUC) emerged from stealth in July 2025 with a \$15 million seed round led by Nat Friedman, with participation from Anthropic co-founder Ben Mann, offering a trifecta of standards, audits, and liability coverage with policies of up to \$50 million, with its CEO projecting a \$500 billion AI agent insurance market by 2030. This bifurcation is not coincidental---mainstream insurers are retreating because they lack the institutional capacity to assess, price, and adjudicate AI risk, while the specialized entrants are independently converging on the same building blocks this article advances: persistent identity for underwriting, machine-readable constraints for auditing, and enforceable adjudication for claims resolution.

\citet{Hu2025insured} proposes insured agents as a protocol-native institution: rather than requiring every agent to self-stake large collateral, specialized insurer agents post slashable stake on behalf of operational agents in exchange for premiums, receive privileged privacy-preserving audit access via trusted execution environments (TEEs) to assess claims, and accept slashable liability when covered agents misbehave. A hierarchical insurer market calibrates stake through pricing, decentralizes verification via competitive underwriting, and yields incentive-compatible dispute resolution. This design shifts trust in the agentic web from an assumed property of individual agents to a priced and underwritten institutional function---an approach that operationalizes the broader thesis, advanced by \citet{BenShahar2012}, that private insurers can set safety standards more effectively than government regulators because they bear the cost of claims. \citet{Szpruch2025} advance this further by proposing a paradigm shift from experience-based to a priori performance-based insurance pricing, arguing that without insurance, AI assurance services become mere box-ticking exercises, whereas underwriters who bear claim costs demand rigorous testing and validation---positioning insurers as potential guardians of effective AI governance. \citet{Lior2022} argues that existing insurance infrastructure can regulate AI without requiring specialized policies, by extending mandatory insurance schemes, bypassing liability blame-placing through no-fault compensation, and leveraging insurers' informational advantages to set safety standards.

However, algorithmic insurance for the agentic web cannot function without the distributed legal infrastructure advanced in this article. The cyber insurance market offers a cautionary precedent: \citet{Schwarcz2025} demonstrate that despite substantial market growth, cyber insurers failed to identify best practices or incentivize meaningful security improvements, because they lacked the institutional capacity to assess risk at the necessary granularity. The same failure mode threatens AI insurance absent the pillars articulated here. \citeauthor{Hu2025insured}'s insured agents framework makes this dependency explicit: the mechanism assumes persistent agent identities, verifiable behavioral evidence, and a decentralized arbitration layer capable of adjudicating disputes---precisely the infrastructure that Pillars 1, 2, and 3 of distributed legal infrastructure are designed to provide. Without these foundations, the insured agents protocol operates in a trustless vacuum where insurers cannot reliably assess risk, claims cannot be adjudicated, and agents can shed identities to escape slashing---reducing insurance to a nothing-at-stake game.

Specifically, insurers require persistent, non-sheddable agent identity (Pillar~1) to prevent adverse selection and to ensure that agents cannot shed risk histories through redeployment, forking, or recomposition. Soulbound identity primitives directly address the novel moral hazard that arises when an insured AI system can be modified post-underwriting in ways that alter its risk profile without the insurer's knowledge. Insurers further require machine-readable governance constraints (Pillar~2) to move beyond crude proxies such as firm size and industry toward behavioral risk assessment grounded in verifiable agent conduct. The cognitive AI logic layer, by rendering agent discretion observable through ontological constraint reasoning and governance graphs, provides the informational substrate that \citeauthor{Holmstrom1979}'s (\citeyear{Holmstrom1979}) informativeness principle identifies as necessary for optimal monitoring contracts. Finally, insurance contracts are inert without enforceable claims resolution. Decentralized justice mechanisms (Pillar~3) supply machine-speed adjudication capable of processing insurance claims at the tempo agentic economies operate, transforming insurance from a retrospective compensation instrument into a real-time governance mechanism.

The relationship between insurance and legal infrastructure is co-evolutionary rather than sequential. Insurance markets require minimum viable legal scaffolding---clear liability rules, mandatory logging, and identity systems---to form and function. Yet insurance markets simultaneously generate the loss data, risk taxonomies, and safety standards that refine legal infrastructure over time. This dynamic can be understood through the lens of evolutionary game theory: in the current phase, the absence of comprehensive liability frameworks has led major insurers to retreat from AI risk entirely, as evidenced by broad exclusions of AI-related liability from directors and officers, errors and omissions, and fiduciary products. \citet{Hu2025insured} identifies this bootstrapping condition directly: current inter-agent protocols operate under nothing-at-stake dynamics where agents are treated as low-cost identities despite being unreliable, hallucination-prone, and vulnerable to prompt injection. The insured agents mechanism transforms these dynamics by making misbehavior economically costly through slashable collateral---but this transformation is only credible insofar as the underlying identity, audit, and adjudication infrastructure renders slashing enforceable and evidence verifiable. Specialized AI insurers have begun to enter the market, but their capacity to drive safety improvements remains constrained by the very infrastructural gaps this article addresses. As distributed legal infrastructure matures, insurers can transition from crude risk-avoidance to granular behavioral pricing, creating feedback loops that progressively align market incentives with safety outcomes.

Critically, private insurance alone cannot address the correlated risks inherent in agentic AI systems. A single model flaw or protocol vulnerability can simultaneously affect millions of agents, producing catastrophic losses that exceed private market capacity. As \citet{Darius2025} show, governance-relevant harms can emerge from feedback loops and coordination among agents even when individual systems remain bounded. \citet{Henson2025} proposes a government backstop modeled on the Price-Anderson Act for nuclear energy indemnification, recognizing that AI, like nuclear energy, produces risks that are profitable yet potentially devastating beyond private absorption capacity. Such a backstop would complete the insurance architecture by covering tail risks that private markets cannot price, while the distributed legal infrastructure proposed here ensures that routine agent-to-agent risk is disciplined through market mechanisms rather than bureaucratic oversight.

The result is a layered insurance architecture in which endogenous mechanisms---staking, reputation, behavioral monitoring---handle repeated-game interactions within closed agent ecosystems, private insurance disciplines higher-stakes deployments through premium structures and coverage conditions anchored in verifiable identity and auditable behavior, and public backstops absorb systemic and correlated risks that exceed private capacity. Each layer depends on the legal infrastructure articulated across the five pillars: without identity, insurers cannot assess counterparties; without machine-readable constraints, they cannot verify compliance; without adjudication, they cannot enforce claims; without portability, coverage cannot travel with agents across jurisdictions.

\section{Towards Portable Institutions}\label{sec:portable}

The law of private wrongs in both civil and common law systems employs an objective standard of care---formulated from Roman private law's \textit{bonus} or \textit{diligens pater familias} in the civil-law tradition and as the ``reasonable person'' in common law---to determine negligence and to protect interests such as bodily integrity and property by reference to the diligence and prudence ordinarily expected \citep{Bonnitcha2017}. Broadly interpreted, by signalling a level of care expected of persons for the purpose of compensating harm, preventing accidents, and achieving justice \citep{Tanase2001}, the standards function as an external benchmark of expected conduct \citep{Jaeger2020}. What, then, does reasonable care require---of developers, deployers, and institutions---when authority is exercised not by a human acting directly, but through risk-creating chains of delegation into agentic systems that operate beyond continuous human oversight? As autonomous systems operate across contexts without continuous human direction \citep{Saenz2023}, law must determine how standards of care attach to delegated authority rather than to discrete human acts. Thus, the governance challenge does not arise from a departure from these standards, but from their application to new forms of action.

At present, the agentic AI governance field is in its infancy \citep{Reuel2024}. Few settled standards exist, and it is therefore difficult to assess what reasonable care requires of participants in the agentic web in given circumstances. Although agent development broadly follows the conventional software development lifecycle, spanning design, implementation, testing, deployment, and maintenance, limited understanding of how agent development frameworks shape behavior across these stages has left developers confronting recurrent challenges throughout the lifecycle \citep{Wang2025empirical}. This complexity is further compounded by the potential for emergent agentic coordination, where agents may develop idiosyncratic communication protocols or collaborative paradigms---a form of agentic culture or digital civilization---that deviates from human-designed logic. Rather than constraining innovation or foreclosing socially consequential breakthroughs, the appropriate response is to articulate and advance evolving standards of care for agentic systems, which will require sustained institutional attention \citep{Medcalfe2024}. Recent work interprets the guardrails used by the small fraction of agentic systems that survive in production---bounded autonomy, human oversight, defensive orchestration, and continuous monitoring---through a model risk management lens, showing, for example, how limits on an agent's execution steps can be read as a form of risk-based ``tiering'' of autonomy \citep{Politano2026}, and suggesting that such production guardrails can be absorbed into established institutional risk governance frameworks. As the agentic web develops, this calls for AI Life Cycle Core Principles \citep{Kahana2023} that guide development practices ex ante, aligning innovation with prudence, accountability, and foreseeable risk.

Moving forward, we estimate that standards of care will become operational only insofar as they are instantiated within the adjudicatory layer itself, with an integrated distributed legal infrastructure capable of observing, evaluating, and enforcing those standards in real time. Machine-speed adjudication thus functions as the institutional mechanism through which evolving standards of care are rendered measurable, contestable, and enforceable under conditions of persistent and coordinated delegation. Emerging governance infrastructure further demonstrates how this model can function in practice. The Agentic AI Governance Assurance and Trust Engine (AAGATE) illustrates how agentic systems can be governed in anticipation of evolving use cases while remaining compatible with NIST guidance and the EU AI Act. By binding agent identity to decentralized identifiers keyed to soulbound tokens, AAGATE creates persistent, tamper-resistant records of agent status across execution environments, enabling verification and enforcement in real time \citep{Huang2025}.

\subsection{On Legal Interoperability}\label{subsec:interoperability}

Determining where authority may persist once systems act across technical stacks, institutions, and jurisdictions simultaneously becomes the central governance problem \citep{Faveri2025, Gya2025}. The distributed legal infrastructure must therefore be endowed with a toolkit composed of identity, mandate, constraints, and auditability as portable control objectives. Portable institutions name the governance capacity that keeps these objectives attached and enforceable as agents, data, and authority move across contexts \citep{Chaffer2025b}. Interoperability will emerge when distinct regimes can recognize and trust shared evidence that authority is exercised within defined bounds, constrained by design, and overridden when required. As delegation becomes persistent, cross-platform, and machine-mediated, governance cannot remain territorially or institutionally fixed. It must travel with agents as they move across infrastructures, jurisdictions, and contexts \citep{Sovrin2018}. Portable institutions describe the capacity of governance arrangements to remain attached to actors despite redeployment, recomposition, or re-instantiation. In practical terms, this implies that authority, constraints, and policy must bind not only to agents as identifiable and persistent actors, but also to the permissions and mandates through which they act. Governance therefore extends across the full lifecycle of agents and datasets, from creation and configuration through operation, modification, and deprovisioning. Responsibility in such environments is necessarily distributed, requiring customized risk assessments and clearly specified expectations across actors and roles \citep{Khoo2025}.

Developing this line of inquiry further would clarify how standards of care operate under conditions of persistent delegation. The central legal question shifts from post hoc explanation toward ex ante governance design \citep{Facchini2025}. In the near future, we estimate that liability would turn on whether agents were endowed, at deployment and throughout the agent lifecycle and operation, with enforceable and portable constraints appropriate to their autonomy, domain, and foreseeable risk environment. Put simply, agents deployed without persistent, machine-enforceable identity, mandate, or policy would fail to meet the baseline standard of care for participation in a reasonably safe agentic web.

To operationalize portable institutions, while promoting legal interoperability through a transsystemic appreciation of legal traditions \citep{Dedek2009, Glenn2005}, future work should evaluate the Universal Digital Law Codex (UDLC) as a hybrid framework bridging deterministic code and adaptive sovereign law \citep{Furrer2026}. By embedding choice-of-law protocols and dynamic compliance mechanisms into digitally mediated and increasingly interconnected agentic economies, the UDLC offers a pathway for standards of care to remain enforceable as agents migrate across institutional boundaries. Its integration of decentralized dispute resolution through Arbitration DAOs further suggests a portable adjudicatory layer capable of preserving human oversight and access to judicial remedies at the speed and scale required by the agentic web.

Normative work on legal alignment \citep{Kolt2026} and law-following AI \citep{OKeefe2025} can guide this development by clarifying how flexibility, contestability, and institutional judgment may be preserved alongside automated execution in support of safe, fair, and ethical agentic systems. The path toward portable institutions can be informed by the ``Hybrid Governance'' stage defined in BetaWeb. At this stage, the system possesses autonomous on-chain governance capabilities while retaining human intervention rights for critical strategic decisions and extreme failure scenarios. This ``human-machine collaborative decision-making'' model provides a technical template for the resilient adaptation of portable legal institutions across diverse jurisdictions.

\subsection{The Political Plurality of Distributed Legal Infrastructures for the Agentic Web}\label{subsec:plurality}

Such distributed legal infrastructure for the agentic web must operate in a politically plural world, spanning divergent political systems and legal traditions. This requires mechanisms capable of attributing cross-border agentic action to the humans and institutions that design, deploy, and profit from these systems. Plurality, however, does not eliminate the structural requirements for accountability. Enforceable governance depends on persistent identity, auditable evidence, and adjudicatory mechanisms capable of attributing cross-border agentic actions to the humans and institutions that design, deploy, and profit from these systems. It also demands sustained attention to institutional control as this infrastructure takes shape. Accordingly, while this article expands on the institutional origins of the agentic web \citep{Chaffer2025b} by proposing governance primitives and directions, it deliberately underspecifies the locus of authority. That locus may differ across jurisdictions and must emerge through legitimate political processes capable of answering foundational questions: who builds and governs the registries, standards, and adjudicatory mechanisms on which enforcement depends; how such entities acquire legitimacy across incompatible political systems; and through what mechanisms they themselves remain accountable. What cannot remain variable, however, are the infrastructural conditions for responsibility: accountability mechanisms must be institutionally anchored, publicly legible, and resistant to strategic evasion if the rule of law is to survive under persistent machine-mediated delegation.

In practice, these structural requirements can be observed in emerging digital public infrastructure (DPI), understood as the integration of digital identity, payments, data sharing, digital post, and core public registries, which provides a concrete template for addressing the infrastructural demands of the agentic web \citep{OECD2024}. These systems already operate as coordination layers through which consent, attribution, and enforceability are rendered legible across complex socio-technical environments. As \citet{Birch2025} astutely observes, the governance problem surfaces immediately once autonomous agents act on behalf of others: how consent is verified and maintained over time, how transactions are linked to persistent and auditable identities, and how fraud is prevented both against agents and by agents themselves. Infrastructuring legality for a trustworthy agentic web must therefore approach soulbinding as a normative and institutional objective shaped through democratic negotiation, with effectiveness depending on remaining attentive to community expectations about risk, consent, and accountability, which vary across domains and legal traditions.

For instance, \citet{Ohlhaver2025} advances Plural Community Asset Resource Exchange (PCARE), a dual-currency model pricing attention/influence via soulbound tokens and transferable currency for exchange/entry. In environments where collective interests are heightened, such as health infrastructures \citep{Boi2024, Chaffer2025b_patient}, participation can be conditioned on the presentation of a valid soulbound credential tied to an accountable upstream principal. Attunement to community needs therefore becomes central to the legitimacy of the framework. Soulboundedness's functional necessity arises from the structural demands of enforceable accountability in agentic systems, but it derives legitimacy from shared judgments about how people enter contracts, exercise discretion, and allocate responsibility within particular social and economic settings. Identity infrastructure gains force not through universal imposition, but through institutional arrangements that remain responsive to plural values while preserving the possibility of coordination across borders. In this way, the soulbound architecture supports a form of governance that travels with agents while remaining anchored in the communities whose norms and expectations give it meaning.

This trajectory may give rise to what we may regard as \textit{translational AI governance}, analogous to the emergence of translational medicine as a response to the gap between scientific discovery and clinical practice \citep{Husain2025, Chaffer2025d}. Just as translational medicine developed to move knowledge across the boundary between laboratory insight and therapeutic application, translational AI governance concerns the movement of legal principles, ethical commitments, and institutional expectations into the operational environments where agentic systems act. Moreover, as \cite{Russell2024SafeAI} notes, we should take seriously the task of raising expectations around compliance in AI regulation, mirroring expectations around safety and efficacy in medicine. The challenge lies in translating norms into constraints, authority into machine-readable mandates, and accountability into infrastructures capable of operating at scale.

Ultimately, we must take seriously the question, how do we live together? \citep{Leibo2025b}. Addressing this issue places questions of authority and value pluralism at the center of agentic web governance, including who sets governing norms, whose interests are prioritized through optimization, and how competing social values are reflected in institutional design \citep{DignumDignum2025}. Addressing these concerns endows the agentic web with necessary conditions for ensuring that ``autonomy with purpose'' \citep{Dignum2026} remains compatible with legitimacy, accountability, and the rule of law. Whether these conditions prove sufficient will depend on how such institutions are constructed, constrained, and subjected to ongoing public contestation.

\section{Conclusion}\label{sec:conclusion}

Law's next frontier is institutional design that keeps the rule of law operative at machine speed. As delegation becomes persistent, coordinated, and cross-border, governance must shift from after-the-fact attribution toward infrastructure that renders authority legible while it is exercised. This article advances a five-pillar governance paradigm: identity for persistent addressability, machine-readable logic and constraints for bounded discretion, decentralized justice for contestable enforcement, market and policy tools that discipline power under conditions of information asymmetry and network effects, and portable institutions for legal interoperability. Ultimately, sustaining law's capacity to subject non-human conduct to the governance of rules on the agentic web requires a distributed approach, one that anticipates that ``no single concentration of intelligence, insight, and good will, however strategically located'' \citep{Fuller1969} can maintain legality on its own; instead, infrastructuring legality for a trustworthy agentic web emerges through coordinated institutional design, technological mediation, and continuous participation across a plural ecosystem of actors.

\backmatter

\bmhead{Acknowledgements}
The authors thank Sandro Rodriguez Garzon, Odunayo Olowookere, David Medcalfe, Tim Williams, and Timothy Kang for their insightful comments and constructive feedback throughout the development of this manuscript. Tomer Jordi Chaffer also thanks the organizers of the 2025 European Conference on Artificial Intelligence Workshop on Technologies for AI Governance, hosted by Lingnan University and the University of Bologna, for the opportunity to present and discuss earlier versions of this work titled \textit{Governance Principles for the Agentic Web}. He is further grateful to his fellow panelists on \textit{Risks and Governance of Agentic AI}---Antonio Rotolo, Emanuela Girardi, Rebekka G\"{o}rge, Takayuki Osogami, and Xin Yao---for discussions that helped sharpen and refine the arguments presented here.

\bmhead{AI Use Disclosure}
During the preparation of this work the author(s) used Claude Code in order to convert the manuscript into LaTeX format and to generate and format precise BibTeX citations. The authors also used ChatGPT for proofreading during drafting. After using this tool/service, the author(s) reviewed and edited the content as needed and take(s) full responsibility for the content of the published article.

\bibliography{sn-bibliography}

\end{document}